\documentclass[letterpaper, 10 pt, conference]{ieeeconf}  %
\IEEEoverridecommandlockouts                              %
 \usepackage[bookmarks=true]{hyperref}
\usepackage{amsmath}
 \usepackage{hyperref}
\usepackage{graphicx}
\usepackage{subfigure}
\usepackage{float}
\usepackage{capt-of}
\usepackage{amssymb}
\usepackage{rotating}
\usepackage{multirow}
\usepackage{bm}
\usepackage{hhline}

\usepackage{graphicx}
\usepackage{setspace}
\usepackage{amsmath}
\usepackage{bm}
\usepackage{subfigure}
\usepackage{tikz}
\usepackage{amssymb}
\usepackage{pifont}
\newcommand{\cmark}{\ding{51}}%

\usepackage{flushend}

\usepackage{caption}
\DeclareMathOperator*{\argmin}{arg\,min}

\title{\LARGE \bf
RGBDTAM: A Cost-Effective and Accurate RGB-D Tracking and Mapping System}

\author{Alejo Concha and Javier Civera
\thanks{Alejo Concha and Javier Civera are with I3A, Universidad de Zaragoza, Spain
        {\tt\small {alejocb,jcivera}@unizar.es}}%
}

\begin{document}

\maketitle
\thispagestyle{empty}
\pagestyle{empty}

\begin{abstract}
Simultaneous Localization and Mapping using RGB-D cameras has been a fertile research topic in the latest decade, due to the suitability of such sensors for indoor robotics. In this paper we propose a direct RGB-D SLAM algorithm with state-of-the-art accuracy and robustness at a los cost. Our experiments in the RGB-D TUM dataset \cite{sturm2012benchmark} effectively show a better accuracy and robustness in CPU real time than direct RGB-D SLAM systems that make use of the GPU.

The key ingredients of our approach are mainly two. Firstly, the combination of a semi-dense photometric and dense geometric error for the pose tracking (see Figure \ref{fig:init_figure}), which we demonstrate to be the most accurate alternative. And secondly, a model of the multi-view constraints and their errors in the mapping and tracking threads, which adds extra information over other approaches. We release the open-source implementation of our approach\footnote{Our code can be found in this link \url{https://github.com/alejocb/rgbdtam}}. The reader is referred to a video with our results \footnote{\url{https://www.youtube.com/watch?v=sc-hqtJtHD4}} for a more illustrative visualization of its performance.

\end{abstract}

\begin{figure}[ht!]

\centering
\subfigure[Sample frame]{
\centering
\includegraphics[width=0.22\textwidth,height=0.17\textwidth]{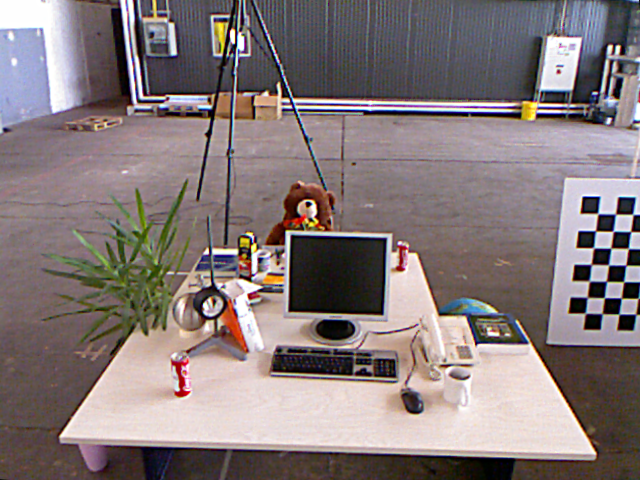}
\label{fig:frame}
} 
\subfigure[Sample frame and map projection]{
\centering
\includegraphics[width=0.22\textwidth,height=0.17\textwidth]{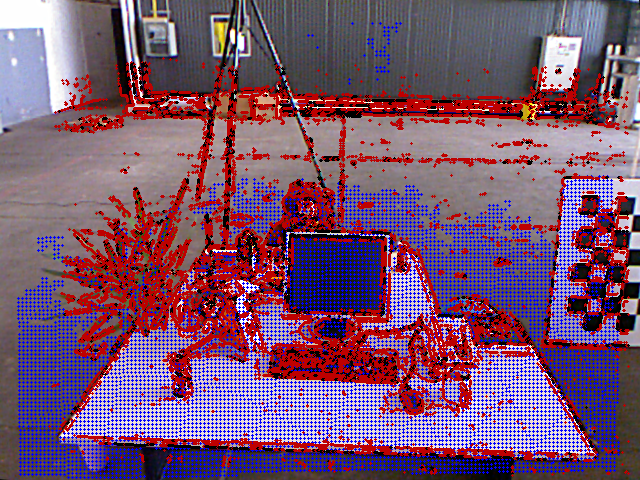}
\label{fig:framerepmap}
}
\subfigure[3D map after back-projecting the depth maps from every keyframe.]{
\centering
\includegraphics[width=0.480\textwidth,height=0.32\textwidth]{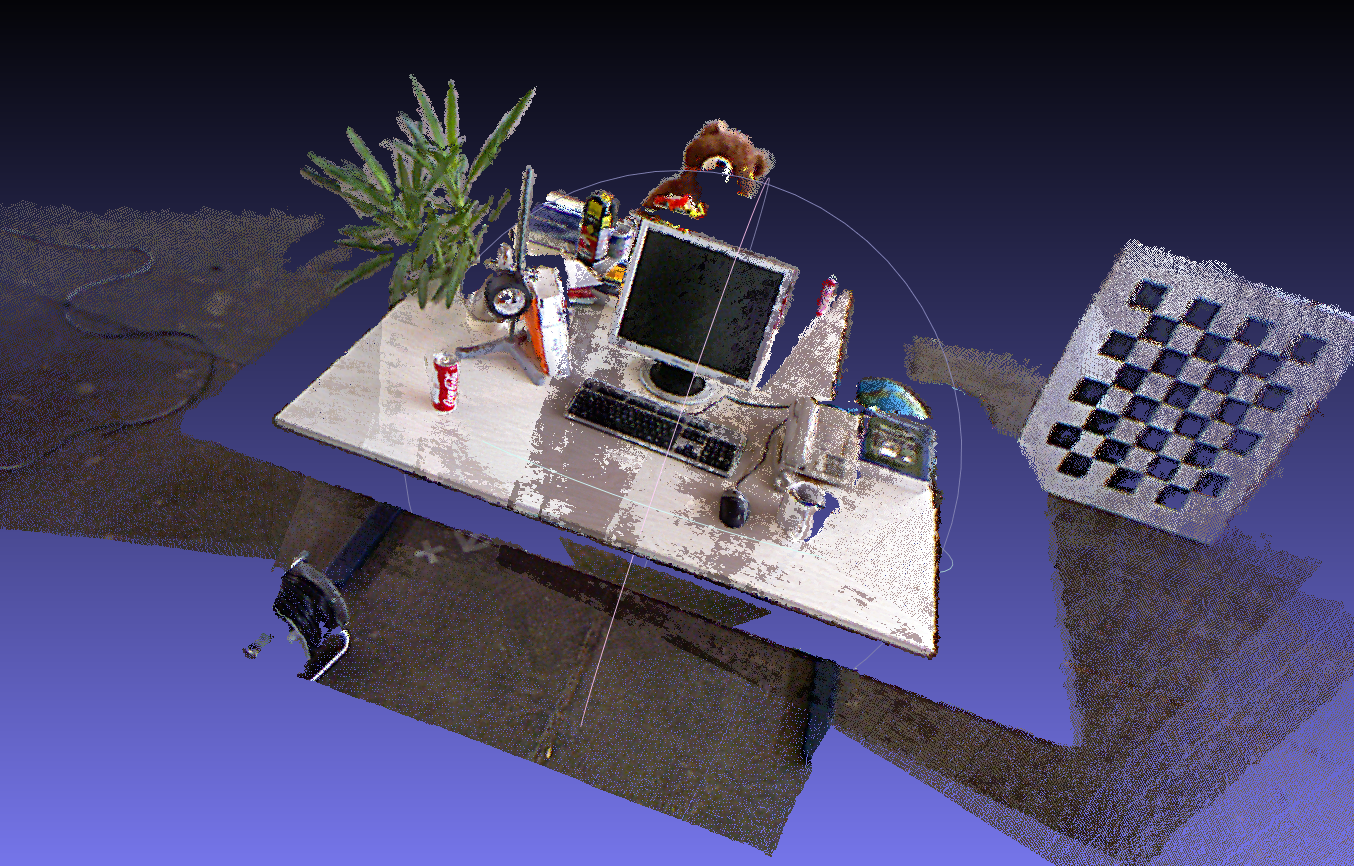}
\label{fig:map}
}
\caption{\subref{fig:frame} Sample frame for one of our experiments. \subref{fig:framerepmap} Same frame with the reprojected map in red and blue. We minimize the photometric error for red points and the geometric error for blue points. Note that distant points are mostly red due to the range limit of the depth sensor. Such points were mapped using multi-view RGB-only constraints. \subref{fig:map} 3D map, composed of the back-projected (non-fused) point clouds from every keyframe.}
\label{fig:init_figure}
\end{figure}

\section{Introduction}
\label{sec:intro}

The availability of affordable and accurate RGB-D cameras has caused a profound impact in mobile robotics. Currently, the research lines based on such technology are as varied as object recognition \cite{lai2011large}, scene recognition and understanding \cite{gupta2015indoor,Song_2015_CVPR}, person detection \cite{spinello2011people} or human-robot interfaces \cite{suarez2012hand}. 

RGB-D sensors have been used also for Visual Odometry (VO) --i.e., the estimation of the incremental motion of the camera from the sensor content-- and SLAM --acronym for Simultaneous Localization and Mapping, aiming at estimating globally consistent scene models in addition to camera ego-motion.
Again, the rationale is the same: RGB-D cameras are perfectly suited to indoor robotics, offering accurate, dense and fully observable measurements within a range at a low cost. Achieving the same accuracy in dense 3D reconstructions from RGB-only sequences is still a challenge \cite{concha2015incorporating}.

RGB-D cameras, however, have several limitations. One of the most relevant is that they cannot operate under direct sunlight. Also, they have a minimum and maximum depth range, and their depth measurements are noisy for absorbent and reflective surfaces. 
In principle, these limitations should have a small effect in the specific applications of VO and SLAM, as they can still use multi-view constraints to estimate the ego-motion and the map. 
However, the state-of-the-art \emph{direct} RGB-D SLAM systems mostly use the depth image constraints and do not fully exploit the information from multiple RGB views.

In this paper we present a direct RGB-D SLAM system that fuses multi-view and depth information. Such fusion extends the range of the maps from the typical few-meters one in RGB-D sensors to potentially infinity. 
Figure \ref{fig:init_figure} illustrates this addition of distant, multi-view points to the map.

Our second contribution is a thorough analysis of the photometric and geometric residual combination. In our experiments, a semi-dense photometric and dense geometric residual has the highest accuracy and robustness. 
Our experimental results in a public dataset shows that RGBDTAM outperforms the state of the art in direct RGB-D SLAM. 

The intuition of the above is, high-gradient pixels are the most informative for multi-view estimation. If the photometric multi-view residual is dense and most of it is composed of low-texture pixels, it is dominated by the noise and hence the estimation is of low accuracy. 

In the case of the geometric error, all the pixels have a high signal/noise ratio. There are some degenerated cases, though, where some degrees of freedom are not constrained, and those justify the combination of both residuals. As they are complementary, the minimization of both errors achieves the best performance. The photometric error is useless in texture-less scenarios, and the geometric one is useless in structure-less scenarios.

The rest of the paper is organized as follows. Section \ref{sec:related} describes the related work. Section \ref{sec:overview} gives an overview of the full RGB-D SLAM system. Section \ref{sec:tracking} details the tracking thread of our SLAM system, section \ref{sec:mapping} the local mapping thread, and section \ref{sec:loopmapreuse} the global mapping and loop closure algorithms we use in our system. Finally, sections \ref{sec:experiments} and \ref{sec:conclusions} show the experimental results and conclusions.

\section{Related work}
\label{sec:related}


One of the first approaches for direct RGB-D odometry is KinectFusion \cite{newcombe2011kinectfusion}, which uses only the depth channel D to estimate the odometry and a dense map and discards the RGB information. As its main limitations, it is restricted to small workspaces and will probably fail if the scene does not contain enough geometric structure.

Kintinuous \cite{whelan2012kintinuous} builds on KinectFusion and uses a rolling cyclical buffer that shifts the volume as the camera is moving, hence not being restricted to small workspaces. It also includes loop closing and pose graph optimization for global consistency.

\cite{steinbrucker2011real} is one of the first approaches that proposes to minimize the photometric error between the current frame and a past frame.

DVO SLAM \cite{kerl2013dense,kerl2013robust} models the map as a pose graph. The constraints between keyframes are set from the tracking thread, which is based on dense photometric and geometric error minimization. This system also achieves CPU real time but not at full resolution ($640\times480$ pixels). 

\cite{klose2013efficient} shows and compares three different alignment strategies for direct tracking, namely the forward-compositional, the inverse-compositional and the efficient second-order minimization approach. In this paper we use the inverse composition, as it is the most efficient of the three.

\cite{endres20143} estimates the relative motion between frames by a least-squares optimization that minimizes the 3D geometric error between corresponding RGB salient points. \cite{henry2012rgb} uses the alignment obtained from feature matching as a seed for a joint optimization of the RGB-D point clouds. In both cases these relative transformations form the edges of a pose graph that are optimized using $\bf g^2 o$ \cite{grisetti2011g2o} and TORO respectively.

Regarding the weighting of the geometric and the photometric error \cite{tykkala2011direct}, \cite{damen2012egocentric}, \cite{meilland2013unifying} and \cite{whelan2013robust} scale the errors with a heuristic constant. \cite{kerl2013dense} weights both contributions according to their respective covariances. \cite{meilland2013super} scale each depth error using its squared inverse depth. 

\cite{gutierrez2015inverse} proposes to use the inverse depth in the minimization of the geometric reprojection error. We evaluate this parametrization in our system.

ElasticFusion \cite{whelan2016elasticfusion} is one of the most recent works and the RGB-D SLAM state of the art in terms of accuracy. The tracking thread uses ICP and dense photometric reprojection error. It achieves global consistency in a map-centric manner by a non-rigid deformation of the map structure, instead of using a more standard pose-centric graph optimization.

Some of these approaches incorporate the multi-view constrains in the tracking thread by weighting the errors with the standard deviation of the depth/inverse depth but they lack a multi-view model in the mapping thread. In our approach we also use multi-view constraints in the mapping. We show in our results that a semi-dense photometric error improves the accuracy and the efficiency of the estimation.  \cite{engel2014lsd,engel2017direct} and \cite{forster2014svo} have shown the effectiveness of a semi-dense or sparse photometric error in the optimization of the camera pose, but in a monocular setting. 

Table \ref{table:systemscomparisontable} details the use of multi-view/depth information and semi-dense/dense residuals for several direct RGB-D SLAM methods in the literature. Notice that our approach is the first one using semi-dense RGB residuals and multi-view constraints for mapping in direct RGB-D SLAM. The reader is referred to section \ref{sec:experiments} for the analysis showing that this combination is the best performing one.

It is worth remarking that some \emph{feature-based} RGB-D SLAM methods (e.g., \cite{henry2012rgb, endres20143, mur2017orb}) use multi-view constraints in the mapping thread and run in real time in a standard CPU. Our contribution and analysis is, however, focused on \emph{direct} methods.

\begin{table*}[ht!]
	\centering
	\begin{tabular}{ | c | c | c | c | c | c | c | c | c | c | c | }
    
      \multicolumn{2}{c}{}  & \multicolumn{2}{c}{\bf } \\
        
		\cline{3-4}
        \hline 
			&   \multicolumn{2}{|c|}{Tracking} &  \multicolumn{2}{|c|}{Mapping}		&  \multicolumn{2}{|c|} {RGB-tracking}  & \multicolumn{2}{|c|} {D-tracking}  \\ \hline  \hline 
		 
       	&D&MV & D & MV & Dense& Semi-dense & Dense& Semi-dense   \\ \hline

       
	 	Newcombe et al. 2011 \cite{newcombe2011kinectfusion} &\cmark &   & \cmark &   &    & &\cmark & \\ \hline
	 	Whelan et al. 2012 \cite{whelan2012kintinuous} &\cmark &  &\cmark& &  \cmark &  &\cmark &  \\ \hline
        
        
	    Kerl et al. 2013 \cite{kerl2013dense} &\cmark &   & \cmark & &  \cmark &  &\cmark &\\ \hline
    	Meilland \& Comport 2013 \cite{meilland2013super} &\cmark    & \cmark & \cmark& &  \cmark &  &\cmark &\\ \hline
    	Gutierrez et al. 2015 \cite{gutierrez2015inverse} &\cmark    & \cmark & \cmark& &  \cmark & &\cmark & \\ \hline
	    Whelan et al. 2016 \cite{whelan2016elasticfusion} &\cmark  & &\cmark & &  \cmark &  &\cmark &\\ \hline
        Jaimez et al. 2017 \cite{jaimez2017fast} &\cmark  & \cmark & & &  \cmark &  &\cmark &\\ \hline
		RGBDTAM & \cmark   & \cmark  & \cmark & \cmark &  &  \cmark  &  \cmark &\\ \hline
		
 		\hline
	\end{tabular}
	\caption{State-of-the-art tracking-and-mapping RGB-D VO/SLAM systems; and their use of depth/multi-view constraints and dense/semi-dense residuals. D stands for depth and MV stands for multi-view.}
	\label{table:systemscomparisontable}
\end{table*}

\section{Notation}
\label{sec:overview}

We follow the standard approach of Parallel Tracking and Mapping, first proposed in \cite{Klein:Murray:ISMAR2007}, and divide our algorithm into two threads. 

The mapping thread estimates a scene map $\mathcal{M}$ from a set of $m$ selected keyframes $\{\mathcal{K}_1, \hdots, \mathcal{K}_j, \hdots, \mathcal{K}_m\}$. Each keyframe  $\mathcal{K}_j=\{{T}_w^j, P^j\}$ is modeled with its pose ${T}_w^j$ in a world frame $w$ and its associated point cloud $P_w^j = \{p_w^1, \hdots, p_w^i, \hdots, p_w^n\}$ where each point $p_w^i$ contains photometric and geometric information. 

The tracking thread estimates the pose of the current frame by minimizing the geometric and photometric reprojection error of its associated point cloud with respect to a previous keyframe. If the scene is revisited the reference keyframe is selected using the method in section \ref{sec:mapreuse}. If the camera is moving through unexplored areas, new keyframes are created based on the camera motion and the overlap with the current point cloud.

\section{Robust RGB-D Tracking}
\label{sec:tracking}

For the camera motion estimation we minimize a functional which is composed of two terms --the photometric error $r_{ph}$ and the geometric error $r_g$--. $r_{ph}$ and  $r_{g}$ will be defined in the following subsections.

\begin{equation}
\label{eq:functional}
\{\hat{T},\hat{a},\hat{b}\} = \argmin_{T,a,b} r_{ph} + \lambda r_g.
\end{equation}

$a$ and $b$ are the gain and brightness of the current image and $\hat{T}$ is the estimated incremental motion of the current camera pose. $\lambda$ is a learned constant weighting the photometric and geometric terms. Notice that we only optimize $T$, $a$ and $b$, therefore we keep the point cloud fixed --for efficiency reasons-- and do not optimize jointly the poses and the points. We use a minimal parametrization for the camera pose, the rotation ${R}$ is mapped into the tangent space $\mathfrak{so}(3)$ of the 
rotation group SO(3) at the identity. Therefore the increments --the angular increment $\delta{\omega}$ and the increment for the translation  $\delta{t}$-- are defined as follows:
\begin{equation} 
{{T}}  = \begin{bmatrix} \exp_{\text{SO(3)}}(\delta {\omega}) & \delta{t}\\ {0}_{1\times 3} & 1\end{bmatrix}.
\end{equation}
\normalsize

We estimate the transformation ${T}_w^f$ from the current camera frame $f$ to the global reference frame $w$ using Gauss-Newton optimization and the inverse compositional approach \cite{baker2004lucas} in equation \ref{eq:functional}.

The update for the current camera pose ${T}_w^f$ is as follows 
\small
\begin{equation}
\label{eq:update1}
{T}_w^f \leftarrow {T}_{w}^f  \hat{{T}}^{-1}.
\end{equation}
\normalsize

\subsection{Photometric error ($r_{ph}$)}
\label{sec:photometric}

We minimize the photometric error only for those pixels belonging to Canny edges \cite{Canny:PAMI86}. Their inverse depth is estimated using the mapping method described in section \ref{sec:mapping}.


The photometric error for the tracking thread is as follows:

\begin{equation}
\label{eq:residual}
r_{ph} = \sum_{i=1}^{n} w_p \left( \frac{\left(I_k(\pi({T}_w^k p_w^i))-aI_f(\pi({T}_w^f  {{T}}^{-1} p_w^i))+b \right)^2}{\sigma_{ph}^2} \right).
\end{equation}

The first term $I_k(\pi(\bm{T}_w^k p_w^i)$ is the intensity of the 3D point $p_w^i$ in a keyframe $I_k$ and the second term $I_f(\pi(\bm{T}_w^f  \hat{{T}}^{-1} p_w^i)))$ is the intensity of the same 3D point in the current frame $I_f$. $\pi()$ is the projection function. Global illumination changes are addressed by estimating $a$ and $b$, which are the gain and brightness of the current frame with respect to the current keyframe. $w_p$ is the Geman-McClure robust cost function, used to remove the influence of occlusions and dynamic objects.

\subsection{Covariance-weighted Geometric error ($r_{g}$)}
\label{sec:geometric}

The second term in equation \ref{eq:functional} is related to the depth measurements. The 3D point cloud is aligned with the current camera and the error between the inverse depth of the points $\frac{1}{{e_z}^T\bm{T}_w^f p_w^i}$ and the measured inverse depth from the depth channel $D_f$ is minimized. 

\begin{equation}
r_g = \sum_{i=1}^{n} w_p \left( \frac{\left(\frac{1}{{e_z}^T{T}_w^f  {{T}}^{-1} p_w^i}-D_f(\pi({T}_w^f  {{T}}^{-1} p_w^i))\right)^2}{\sigma_g^2} \right).
\end{equation}

$w_p$ is again the Geman-McClure robust cost function. $e_z$ is a 3D vector defined as $e_z = [0,0,1]$

Contrary the photometric residual (detailed in section \ref{sec:photometric}), in this case a dense optimization is better than a semi-dense one. 
We homogeneously subsample the pixels used in the geometric error, in order to achieve CPU real-time performance. 
We use four pyramid levels (from $80\times60$ to $640\times480$). For the first level we use all pixels. For the second, third and fourth levels we use one in every two, three and four pixels respectively --horizontally and vertically. 

\paragraph*{Covariance Propagation for Structured Light Cameras}

In order to estimate a value for the standard deviation of the geometric residual $\sigma_g$, we model the depth error of RGB-D cameras as that of the stereo. Our analysis is valid for RGB-D sensors based on structured light patterns, such as the Kinect v1 or the Google Tango. 

Focusing our analysis in the epipolar plane, the stereo depth $z$ only depends on the disparity $d$, the camera focal length $f$ and the baseline $b$

\begin{equation}
\centering
z = \frac{fb}{d} ~.
\end{equation}

For the inverse depth $\rho$

\begin{equation}
\centering
\rho= \frac{d}{fb} ~.
\end{equation}

Assuming a disparity error with standard deviation $\sigma_d$ (and no error for the focal length and the baseline), a first-order propagation gives the following standard deviation for the depth error

\begin{equation}
\centering
\label{eq:sigma_z}
\sigma_z = \frac{\partial z}{\partial d}\sigma_d = \frac{fb}{d^2}\sigma_d = \frac{z^2}{fb}\sigma_d~.
\end{equation}

The inverse depth parametrization \cite{Civera:etal:TR2008} is linearly dependent on the disparity. The first order error propagation gives, for a fixed baseline, a constant uncertainty in the inverse depth.

\begin{equation}
\centering
\label{eq:sigma_rho}
\sigma_{\rho} = \frac{\partial \rho}{\partial d}\sigma_d = \frac{\sigma_d}{fb}~.
\end{equation}


\subsection{Scaling parameters}
\label{sec:scale}

As we combine residuals of different magnitudes, we need to scale them according to their covariances. For the geometric error we propagate its uncertainty using equations \ref{eq:sigma_z} and \ref{eq:sigma_rho}. For the photometric error we use the median absolute deviation of the residuals of the previous frame to extract a robust estimation of the standard deviation.

\begin{equation}
\sigma_{ph}  =  1.482  * \text{median}(r_{ph} - \text{median}(r_{ph})).
\end{equation}

\section{RGB-D Mapping}
\label{sec:mapping}

We add a new keyframe in the map when the percentage of pixels that is visible from the previous keyframe is below a threshold. The mapping thread estimates a semi-dense map as soon as possible, in order to minimize the tracking failure risk.

Every pixel may have up to two sources of information to estimate its inverse depth: The raw depth sensor reading ($\rho_1$) and multi-view geometry ($\rho_2$). 


For the multi-view triangulation we follow an approach similar to \cite{engel2014lsd,Concha:Civera:IROS15}. The inverse depth $\rho_2$ for every high-gradient pixel ${u}^*$ in a keyframe ${I}_j$ is estimated by minimizing its photometric error ${r}^o_{ph}$ with respect to several overlapping views ${I}_o$. 

 
\begin{equation}
\hat{\rho}_2 = \argmin_{\rho_2} {r}_{ph},
\end{equation}

with
 
\begin{equation}
\label{eq:residual_semidense}
{r}_{ph} = \sum_o\left\Vert\left({I}_j\left({s}_{u^*}\right) - {I}_o\left(G\left({s}_{u^*},{T}_w^j,{T}_w^o,\rho\right)\right)\right)\right\Vert_2^2.
\end{equation}

${s}_{u^*}$ are the pixel coordinates of the template (we use a one-dimensional patch, similarly to \cite{engel2014lsd}) around the pixel ${u}^*$ and G is the function that backprojects the template from the new keyframe ${I}_j$ to the 3D world and projects it back to each overlapping  image ${I}_o$.


These two contributions are fused using their uncertainties as follows

\begin{align}
\begin{aligned}
\rho =\frac{\sum_{j=1}^{2} \frac{\rho_j}{\sigma_j^2}} {\sum_{j=1}^{2}  \frac{1}{\sigma_j^2}},   &
& \sigma =\frac{1} {\sum_{j=1}^{2} \frac{1}{\sigma_j^2}}.
\end{aligned}
\end{align}

The uncertainties $\sigma_j$ are estimated using equation \ref{eq:sigma_rho}. Notice that we do not fuse the inverse depth map of the current keyframe with the inverse depth map of the previous keyframes or the 3D model. There is a reason for this. We do not optimize jointly the pose of the keyframes and the 3D point cloud (as it is done in most direct algorithms \cite{newcombe2011dtam,engel2014lsd}). The fusion of different depth maps transfers the errors of each keyframe to the 3D map, resulting in a less accurate localization of the camera.

\section{Loop closure and map reuse}
\label{sec:loopmapreuse}

\subsection{Loop closure}
\label{sec:loopclosing}
The back-end of our algorithm is composed of loop closure detection and pose-graph optimization over the keyframes. We used the open library DBoW2 \cite{GalvezTRO12} for appearance-based loop closure and the vocabulary created by the ORB-SLAM authors \cite{Mur:etal:15a}. The ratio between the best match (a previous keyframe) and a neighboring keyframe of the current keyframe is calculated. If this ratio is higher than a threshold (0.5 in our experiments), the previous keyframe becomes a candidate for loop closing.

Once the candidate has been selected, we search for ORB \cite{rublee2011orb} correspondences in both keyframes and use RANSAC \cite{Fischler1981} to get the 6 DOF transformation between the sparse point clouds. We use the Horn's method \cite{horn1987closed} to calculate this transformation ${T}_j^k$ between keyframes $j$ and $k$. We define a point as an inlier if the reprojection error --taking into account the pyramid level-- is smaller than a threshold. If a minimum number of inliers is found the loop closure is accepted. Once the loop is detected the 6-DoF poses of all the keyframes are refined using pose-graph optimization with the $\bf g^2 o$ library \cite{grisetti2011g2o}. The following functional is minimized:  

\begin{equation} 
\label{eq:posegraph1}
\{ {\hat{T}_w^1}, \hdots, {\hat{T}_w^j}, \hdots, {\hat{T}_w^k} \hdots, {\hat{T}_w^m} \} = \argmin_{\{ {{T}_w^1}, \hdots, {{T}_w^m} \}} \sum_{j,k} r_{j,k}^\top \Lambda_{j,k} r_{j,k} 
\end{equation}

Where $\{ {{T}_w^1}, \hdots, {{T}_w^j}, \hdots, {{T}_w^k} \hdots, {{T}_w^m} \}$ are the poses of the $n$ keyframes in the map. $\Lambda_{j,k}$ is the information matrix, which we set to the identity.  $r_{j,k}$ is the residual for the 
edge $j,k$ which is defined as follows:

\begin{equation} 
\label{eq:posegraph2}
r_{j,k} = \log{\left(T_j^k T_k^w T_w^j\right)}   .
\end{equation}



\subsection{Map reuse}
\label{sec:mapreuse}

Instead of continuously creating new keyframes we adopt a conservative strategy that privileges the use of the already existing ones. Our system looks for overlapping keyframes in a specific area before creating a new one, and in this manner we reduce the accumulated drift. Again, we use DBoW2 \cite{GalvezTRO12} to obtain a list of candidate keyframes imaging the current tracked area. We propose two heuristic rules to discard invalid candidates:

\begin{itemize}
\item{An overlap of at least $80\%$ between the previous keyframe and the current frame is required.}
\item{The photometric and geometric reprojection error are required to be smaller than 3 times the standard deviation of both errors. For the photometric error we set $\sigma_{ph} = 15$};
\end{itemize}

After applying these heuristics to remove loop outliers, we select the oldest candidate keyframe and use it for tracking. Notice that the pose of the old keyframe is taken after the optimization of the pose-graph functional --equation \ref{eq:posegraph1}.

If these heuristics do not hold for any previous keyframe, then we try to close the loop following the approach described in the previous section \ref{sec:loopclosing}. Finally, if these described strategies do not succeed we assume the system is exploring new areas and create a new keyframe.

\section{Experimental Results}
\label{sec:experiments}

For our experimental results we use the publicly available TUM dataset \cite{sturm2012benchmark}. We have done an exhaustive evaluation in all the static sequences of the dataset. We run our code on every sequence $5$ times with different random initialization parameters and report the median of the $5$ trajectory errors. 

\paragraph{\bf Comparison against direct RGB-D SLAM}
\label{sec:elastic}
We compare our approach against ElasticFusion \cite{whelan2016elasticfusion}, a state-of-the-art direct RGB-D SLAM system. We do not include other direct approaches in the evaluation, as \cite{whelan2016elasticfusion} outperforms all the previous baselines. 

Table \ref{table:elasticfusion} shows the trajectory error (RMSE). RGBDTAM shows better accuracy and robustness in most of the sequences. We used the same sequences than \cite{whelan2016elasticfusion}. 
RGBDTAM is in general more robust than ElasticFusion in sequences with poor structure and texture in close objects (sequences fr2 coke, fr2 dishes, fr2 metallic sphere, fr3 cabinet). The semi-dense photometric residual is more robust in these sequences, as textureless areas do not contribute to the optimization.

\begin{table}[ht!]
	\centering
    \small
	\begin{tabular}{ | c || l || c | c |  }

        \multicolumn{2}{c}{}  & \multicolumn{2}{c}{\bf } \\
    
		\cline{3-4}
        \multicolumn{2}{c||}{}  & \multicolumn{2}{|c|}{\bf RMSE [cm]} \\
        \hline 
		{ \bf \#}	& {\bf  Sequence Name}		& \multicolumn{1}{|c|}{\bf  \cite{whelan2016elasticfusion} }& {\bf RGBDTAM}  \\ \hline  \hline 
		
		1&fr1 360  & 10.8 & \bf{10.1} \\ \hline
		2&fr1 desk  &\bf{2.0} & 2.7 \\ \hline
		3&fr1 desk2  & 4.8 &\bf{4.2} \\ \hline
		4&fr1 floor  & -  & - \\ \hline
		5&fr1 plant  & \bf{2.2} & 2.5 \\ \hline
		6&fr1 room  & \bf{6.8} & 15.5 \\ \hline
		7&fr1 rpy  & 2.5 & \bf{2.1} \\ \hline
		8&fr1 teddy  &8.3 &  \bf{8.1} \\ \hline
		9&fr1 xyz & 1.1 & \bf{1.0} \\ \hline
		10&fr2 360 hemisphere  & - & - \\ \hline
		11&fr2 360 kidnap  & - & -\\ \hline
		12&fr2 coke  & - & \bf{6.0}\\ \hline
		13&fr2 desk  &7.1 &\bf{2.7}\\ \hline
		14&fr2 dishes  &- & \bf{3.6}\\ \hline
		15&fr2 large no loop  &- & - \\ \hline
		16&fr2 large with loop  & - & - \\ \hline
		17&fr2 metallic sphere  & - & - \\ \hline
		18&fr2 metallic sphere 2  & - & \bf{5.2} \\ \hline
		19&fr2 pioneer 360  & - &- \\ \hline
		20&fr2 pioneer slam  & - & - \\ \hline
		21&fr2 pioneer slam2  & -& - \\ \hline
		22&fr2 pioneer slam3  & - & - \\ \hline
		23&fr2 rpy  & 1.5 & \bf{0.2} \\ \hline
		24&fr2 xyz  & 1.1 & \bf{0.7} \\ \hline
		25&fr3 cabinet  & - &\bf{5.7}\\ \hline
		26&fr3 large cabinet  &9.9 & \bf{7.0} \\ \hline
		27&fr3 long office household  & \bf{1.7} &2.7 \\ \hline
		28&fr3 nostr. notext. far  & - & - \\ \hline
		29&fr3 nostr. notext. near withloop  &- &- \\ \hline
        30&fr3 nostr. text. far  & 7.4 & \bf{2.6} \\ \hline
		31&fr3 nostr. text. near withloop  & 1.6 &  \bf{1.0} \\ \hline
		32&fr3 str. notext. far  & 3.0 & \bf{1.3}\\ \hline
		33&fr3 str. notext. near &\bf{2.1} & 4.4 \\ \hline
	    34&fr3 str. text. far  & 1.3 & \bf{1.0} \\ \hline
		35&fr3 str. text. near & 1.5 & \bf{1.0}  \\ \hline
		36&fr3 teddy &\bf{4.9} & - \\ \hline

		\hline
	\end{tabular}
	\caption{RMSE for ElasticFusion \cite{whelan2016elasticfusion} and RGBDTAM in the static sequences of \cite{sturm2012benchmark}. Results from \cite{whelan2016elasticfusion} are reported with best per-sequence-parameters; ours are with best per-dataset-parameters.}
	\label{table:elasticfusion}
\end{table}

\normalsize

The higher accuracy, higher robustness and lower cost of RGBDTAM with respect to ElasticFusion is remarkable and deserves further elaboration. ElasticFusion aims to estimate a map-centric global representation by an non-rigid fusion of the RGB-D keyframes; estimating visually appealing maps with an appearance of global consistency. RGBDTAM, by adopting the more traditional pose-centric approach, focuses on the solid probabilistic integration of the most informative data, and presumably this is the reason of its higher accuracy. We believe that the two approaches are complementary and valuable and would like to combine their respective strengths in future work. 

\paragraph{\bf Comparison against feature-based RGB-D SLAM}
\label{sec:orbslam}

We compare RGBDTAM against ORB-SLAM2 \cite{mur2017orb}, the state-of-the-art feature-based RGB-D SLAM system. 

Table \ref{table:orbslam} shows the trajectory error (RMSE) in the TUM dataset. RGBDTAM has worse accuracy in most of the sequences. We have used the same sequences than the original paper \cite{mur2017orb}. We believe the reason for our worse performance is that RGBDTAM alternates between tracking and triangulation, lacking a joint optimization of poses and points in the mapping thread. 
 
\begin{table}[ht!]
	\centering
    \small
	\begin{tabular}{ | c || l || c | c |  }

        \multicolumn{2}{c}{}  & \multicolumn{2}{c}{\bf } \\
    
		\cline{3-4}
        \multicolumn{2}{c||}{}  & \multicolumn{2}{|c|}{\bf RMSE [cm]} \\
        \hline 
		{ \bf \#}	& {\bf  Sequence Name}		& \multicolumn{1}{|c|}{\bf  \cite{mur2017orb} }& {\bf   RGBDTAM}  \\ \hline  \hline 
		
		2&fr1 desk  &\bf{1.6} & 2.7 \\ \hline
		3&fr1 desk2  & \bf{2.2} &4.2 \\ \hline
        6&fr1 room  & \bf{4.8} &15.5 \\ \hline
		13&fr2 desk  &\bf{0.9} &2.7\\ \hline
		24&fr2 xyz  & \bf{0.4} & \bf{0.4} \\ \hline
		27&fr3 long office household  & \bf{1.0} &2.7 \\ \hline
        29&fr3 nstr  & 1.9 &\bf{1.6} \\ \hline

		\hline
	\end{tabular}
	\caption{RMSE for ORBSLAM2 \cite{mur2017orb} and RGBDTAM}
	\label{table:orbslam}
\end{table}

\paragraph{\bf Evaluation of the residual configuration}
\label{sec:evaluation_geo_photo}

Table \ref{table:photodensevssemi} shows a comparison between different configurations in the tracking thread. Observe that minimizing the photometric error for a semi-dense subset of high-gradient pixels (\emph{PS} row in the table) is in general more accurate than using the full image (\emph{PD} row in the table). 
Notice also that a semi-dense approach is more robust. When using a dense approach the camera track was lost in some sequences with very little texture, where the noise of textureless areas in the image and other artifacts (such as reflections) dominated the solution. 

\begin{table*}[ht!]
	\centering
    \small

        \begin{tabular}{  |l | c| c | c| c | c | c| c |c|  c |c | c | c | c | }
        
              \multicolumn{2}{c}{}  & \multicolumn{2}{c}{\bf } \\
		\hline 
		{ \bf \#Seq.}	& {1}& {2} &{3}& {4} & {5}& {6} & {7}& {8} & {9}& {10} & {11}& {12}  		 \\ \hline 
        
{ \tiny{ { \bf PS [RMSE cm].}}}		& \bf{9.3}& {2.7} &{6.4}& {-} & {4.4}& {13.5} & {2.5}& {14.1} & \bf{1.0}& {-} & {-}& {8.8} 	 	 \\ \hline  
 
 { \tiny{ { \bf PD [RMSE cm].}}}		& {12.4}& {5.0} &{8.4}& {-} & {8.2}& {25.0} & {2.3}& {12.6} & \bf{1.0}& {-} & {-}& {-} 	 	 \\ \hline  
  { \tiny{ { \bf GIDS [RMSE cm].}}}		& {12.7}& {3.2} &{6.5}& {-} & {6.9}& \bf{12.5} & {5.4}& {12.5} & {6.3}& {-} & {-}& {-} 	 	 \\ \hline  
   { \tiny{ { \bf GIDD [RMSE cm].}}}		& {13.4}& {3.4} &{6.6}& {-} & {6.1}& {12.9} & \bf{1.7}& {9.5} & {1.3}& {-} & {-}& {-} 	 	 \\ \hline  
{ \tiny{ { \bf GIDD* [RMSE cm].}}}		& {13.2}& {3.0} &{6.3}& {-} & {5.0}& {15.1} & {1.8}& {8.1} & {1.2}& {-} & {-}& {-} 	 	 \\ \hline  
   { \tiny{ { \bf PS + GIDD [RMSE cm].}}}	&	{10.1}& {2.7} & \bf{4.2}&  {-}& \bf{2.5} & {15.5}& {2.1} & {8.1}& \bf{1.0}  & {-}& {-} & \bf{6.0}\\ \hline 
   { \tiny{ { \bf PS + GDD [RMSE cm].}}}	&	{9.4}& \bf{2.5} & {4.3}&  {-}& {2.9} & {16.1}& {2.3} & \bf{8.0}& \bf{1.0}  & {-}& {-} & {7.2} \\ \hline 

\multicolumn{13}{c}{}
\\ 
\multicolumn{13}{c}{}
\\ \hline 

		{ \bf \#Seq.}	& {13}& {14} &{15}& {16} & {17}& {18} & {19}& {20} & {21}& {22} & {23}& {24}  		 \\ \hline            
 { \tiny{ { \bf PS [RMSE cm].}}}		& {2.6}& {3.6} &{-}& {-} & {-}& {9.3} & {-}& {-} & {-}& {-} & \bf{0.2}& {0.5} 	 	 \\ \hline  
 { \tiny{ { \bf PD [RMSE cm].}}}		& {12.1}& {15.7} &{-}& {-} & {-}& {-} & {-}& {-} & {-}& {-} & {0.3}& \bf{0.4} 	 	 \\ \hline  
  { \tiny{ { \bf GIDS [RMSE cm].}}}		& {8.3}& {-} &{-}& {-} & {-}& {-} & {-}& {-} & {-}& {-} & {3.6}& {1.9} 	 	 \\ \hline  
   { \tiny{ { \bf GIDD [RMSE cm].}}}		& {11.9}& {-} &{-}& {-} & {-}& {-} & {-}& {-} & {-}& {-} & {3.3}& {1.8} 	 	 \\ \hline  
   { \tiny{ { \bf GIDD* [RMSE cm].}}}		& {11.7}& {-} &{-}& {-} & {-}& {-} & {-}& {-} & {-}& {-} & {3.2}& {1.7} 	 	 \\ \hline  
    { \tiny{ { \bf PS + GIDD [RMSE cm].}}}		& {2.7}& {3.6} &{-}& {-} & {-}& \bf{5.2} & {-}& {-} & {-}& {-} & \bf{0.2}& {0.7} 				 \\ \hline  
  { \tiny{ { \bf PS + GDD [RMSE cm].}}}		& \bf{2.3}& \bf{3.3} &{-}& {-} & {-}& {5.4} & {-}& {-} & {-}& {-} & \bf{0.2}& {0.6} \\ \hline  
  
\multicolumn{13}{c}{}
\\ 
\multicolumn{13}{c}{}
\\ \hline 

		{ \bf \#Seq.}	& {25}& {26} &{27}& {28} & {29}& {30} & {31}& {32} & {33}& {34} & {35}& {36}  		 \\ \hline            

 { \tiny{ { \bf PS [RMSE cm].}}}		& {-}& \bf{5.3} &\bf{2.7}& {-} & {-}& {3.3} & {1.5}& {7.5} & {-}& {1.2} & {1.3}& {-} 	 	 \\ \hline  
 { \tiny{ { \bf PD [RMSE cm].}}}		& {-}& {-} &{-}& {-} & {-}& {-} & {13.5}& {2.9} & {-}& {2.0} & {1.2}& {-} 	 	 \\ \hline 
  { \tiny{ { \bf GIDS [RMSE cm].}}}		& {9.7}& {-} &{-}& {-} & {-}& {-} & {-}& {4.2} & {3.3}& {3.4} & {6.5}& {-} 	 	 \\ \hline  
   { \tiny{ { \bf GIDD [RMSE cm].}}}		& {-}& {-} &{-}& {-} & {-}& {-} & {-}& {8.0} & {3.0}& {4.0} & {5.1}& {-} 	 	 \\ \hline  
      { \tiny{ { \bf GIDD* [RMSE cm].}}}		& {-}& {-} &{-}& {-} & {-}& {-} & {-}& {7.1} & {2.6}& {3.3} & {6.2}& {-} 	 	 \\ \hline  
    { \tiny{ { \bf PS + GIDD [RMSE cm].}}}
 &\bf{5.7}& {7.0} &\bf {2.7} & {-}& {-} & \bf{2.6}& \bf{1.0} & \bf{1.3}& {4.4} & {1.0}  & \bf{1.0} & {-} 				 \\ \hline  
  { \tiny{ { \bf PS + GDD [RMSE cm].}}}
 &{8.8}& {8.5} &\bf {2.7} & {-}& {-} & {3.5}& {1.1} & {1.4}& \bf{4.0} &\bf {0.9}  & {1.2} & {-} 				 \\ \hline  

 \end{tabular}

	\caption{RMSE for different RGBDTAM configurations. \emph{PS} stands for photometric semi-dense, and \emph{PD} for photometric dense. \emph{GIDD} and \emph{GIDD*} stand for geometric inverse depth dense. \emph{GIDD} only uses a subsample of the points  (homogeneously distributed). \emph{GDD} stands for geometric depth dense. \emph{GIDS} stands for geometric inverse depth semi-dense. The combination of a semi-dense photometric and a dense --subsampled-- geometric is the most accurate and robust.}
	\label{table:photodensevssemi}
\end{table*}

\normalsize

Compare now the errors between a semi-dense and a dense point cloud for the geometric depth error (\emph{GIDS} and \emph{GIDD} rows respectively). Notice that in this case a dense approach is more accurate. 
We homogeneously selected a subset of the points of the geometric point cloud in order to achieve real-time performance. We have observed that this reduction of the dimensionality of the geometric error does not impact the performance of our approach. Notice in table \ref{table:photodensevssemi} that the subsampled version \emph{GIDD} only obtains slightly less accurate results than the fully dense version \emph{GIDD*}. The best configuration in terms of efficiency, accuracy and robustness is the one that fuses the semi-dense photometric error and a subsampled of the dense geometric error in the optimization (\emph{PS+GIDD} row).  

\paragraph{\bf Depth vs inverse depth in the geometric reprojection error}
 
The last two rows in Table \ref{table:photodensevssemi} reports the comparison between the the inverse depth and the depth in the minimization of the geometric error for the best performing configuration of the tracking thread.  

We obtained slightly better results for the inverse depth case, but the difference is small due to the range limits of RGB-D sensors. The inverse depth is particularly useful for distant points, for which a depth sensor does not measure its depth.
 
 \paragraph{\bf Computational time}
\label{sec:time}
All the experiments 
were run in CPU real-time (the average time being $35.86 ms$ per frame) on a laptop with a 3.5 GHz Intel Core i7-3770K processor and 8.0 GB of RAM memory. Notice that ElasticFusion needs GPU processing and hence, being the comparison difficult, their cost will be presumably higher than ours. 

For the semi-dense (contour based) photometric error we track up to $8K$ points per keyframe. For the geometric error, we homogeneously subsample it as explained in section \ref{sec:mapping}. This configuration is the reported in tables \ref{table:elasticfusion}, \ref{table:orbslam} and \ref{table:photodensevssemi}.

 \paragraph{\bf Failure modes}
 \label{sec:failure}

Notice in Table \ref{table:elasticfusion} that RGDTAM failed in 13 sequences out of 36. 
For the first 9 sequences there were missing frames that led to tracking failure in our system. In sequence 17 a relatively big and slow-moving dynamic object was not rejected as outlier by the robust cost function. For the next two sequences the scene did not contain structure nor texture, as the camera was moving on top of a textureless floor. Both the geometric and photometric errors were uninformative. For the last case (sequence number 36), the camera came very close to an object and performed a pure rotation. The object was closest than the minimum depth range and the multi-view mapping was not able to estimate a map without parallax. 	

 \paragraph{\bf Qualitative results.}

\begin{figure*}[ht!]
\centering
\subfigure[Sequence fr2 rpy]{
\centering
\includegraphics[width=0.48\textwidth,height=0.35\textwidth]{fr1_rpy.png}
\label{fig:qualitativea}
}
\subfigure[Sequence fr3 household long office]{
\centering
\includegraphics[width=0.48\textwidth,height=0.35\textwidth]{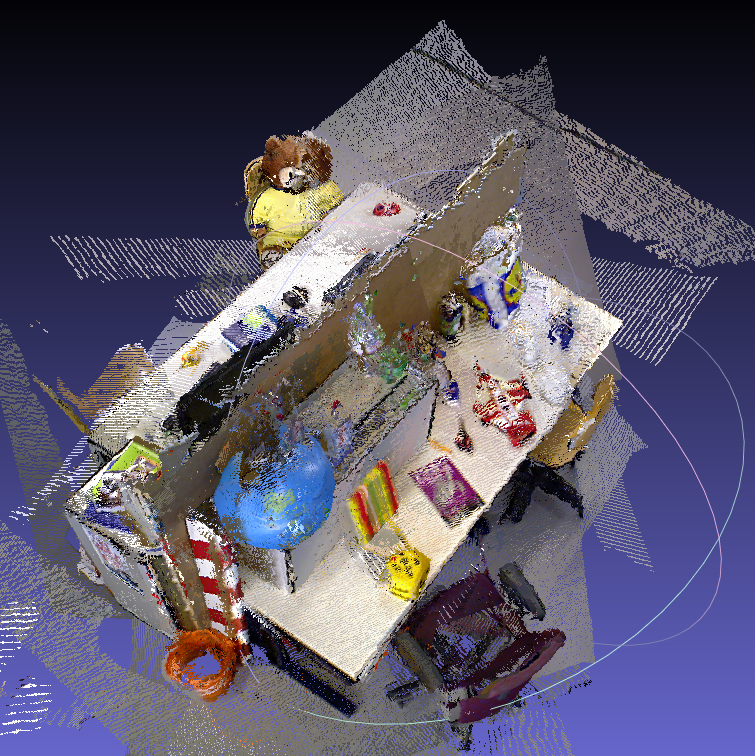}
\label{fig:qualitativeb}
}
\subfigure[Sequence fr3 no structure texture near with loop]{
\centering
\includegraphics[width=0.48\textwidth,height=0.35\textwidth]{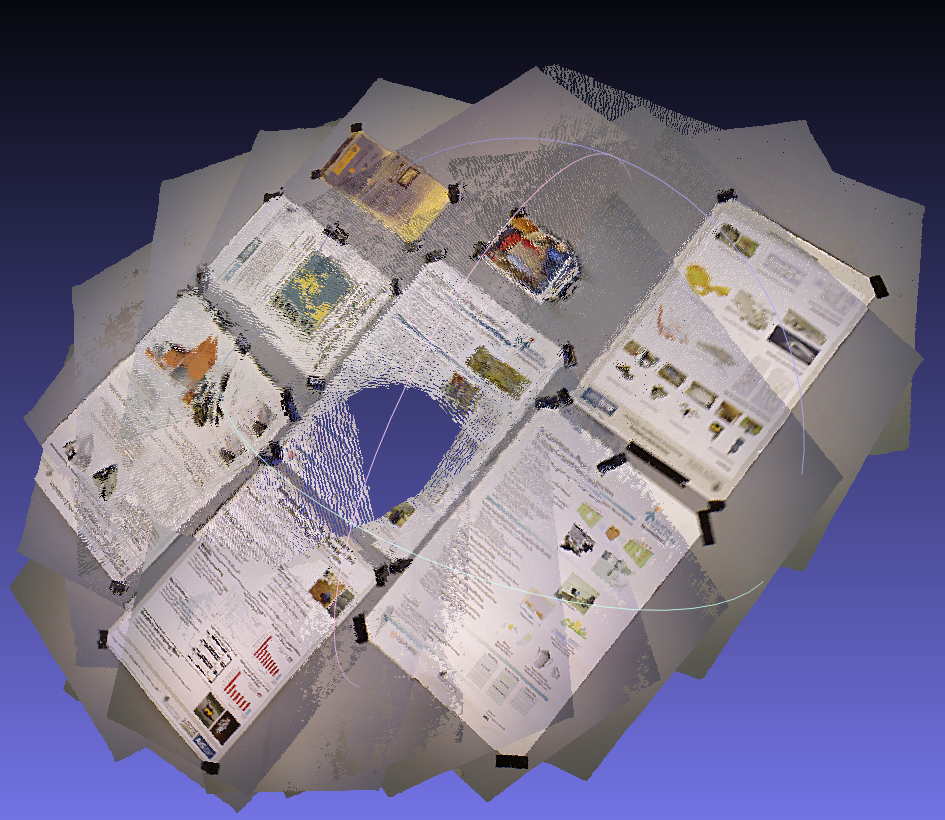}
\label{fig:qualitativec}
}
\subfigure[Sequence fr3 structure no texture far]{
\centering
\includegraphics[width=0.48\textwidth,height=0.35\textwidth]{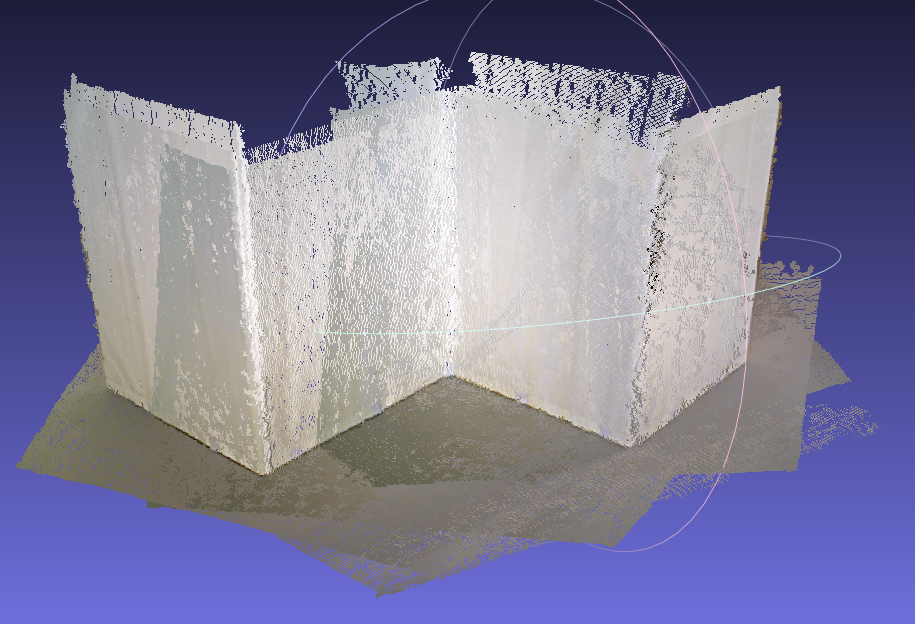}
\label{fig:qualitatived}
}

\caption{Qualitative results. Depth maps are not fused. They are back projected from every keyframe.}
\label{fig:qualitative}
\end{figure*}

Figure \ref{fig:qualitative} shows several 3D maps obtained by our system in the TUM dataset. Notice the high accuracy of the map, even when we do not fuse the point clouds from different keyframes. Notice also that the 3D reconstruction in Fig. \ref{fig:qualitativec} corresponds to a structure-less but textured scene, and hence accurate thanks to the photometric part of the residual. On the other side, the 3D reconstruction in Fig. \ref{fig:qualitatived} is of a textureless scene of rich structure, and hence only possible thanks to the geometric part of the residual. See table \ref{table:photodensevssemi} for quantitative results in these sequences. 

\section{Conclusions}
\label{sec:conclusions}

In this paper we have presented a direct RGB-D SLAM system with loop closure and map reuse capabilities. Our main contribution is the integration of RGB multi-view constraints in both the tracking and mapping thread. Such multi-view constraints increase the accuracy of the estimation due to two factors. First, the addition of distant points, out of the RGB-D sensor range, to the map. And second, the extra accuracy gained in high-parallax configurations. 

We have compared different settings for the photometric and the geometric residual in the tracking thread, concluding that a combination of a semi-dense photometric error and a dense geometric error is the best combination in terms of accuracy and robustness. We have evaluated the minimization of the depth and inverse depth in the geometric error. The inverse depth parametrization is slightly more accurate in our results. Finally, we have shown that our approach outperforms the state of the art on direct RGB-D SLAM systems in terms of trajectory accuracy. Our system is also amongst the ones with the lowest cost, running in real time on a standard CPU.

\section*{Acknowledgments}
This research has been partially funded by the Spanish government (project DPI2015-67275) and the Arag\'on regional government (Grupo DGA T04-FSE). 


\bibliographystyle{plain}
\bibliography{alejo}

\end{document}